\documentclass[conference,comsoc]{IEEEtran}
\usepackage[T1]{fontenc}
\usepackage{amsmath}
\usepackage[cmintegrals]{newtxmath}
\usepackage{url}
\usepackage{soul,color,graphicx,float,epstopdf}
\usepackage{tablefootnote,textcomp,gensymb}
\usepackage{eurosym,booktabs,multirow}
\usepackage[caption=false]{subfig}
\usepackage{amsfonts} 
\usepackage{algorithmic}
\usepackage{array}
\usepackage{stfloats}
\usepackage{float}
\usepackage{mathtools}
\usepackage{graphicx}
\usepackage{pdfpages}
\usepackage{subfig} 
\usepackage{comment}
\usepackage{balance}
\usepackage{xcolor}
\usepackage{comment}
\usepackage{algorithm}
\usepackage[normalem]{ulem}
\usepackage[implicit=false]{hyperref}
\usepackage{enumitem}
\definecolor{Orange}{rgb}{1,0.5,0}
\newcommand{\todo}[1]{\textsf{\textbf{\textcolor{Orange}{[#1]}}}}

\begin{document}
\title{Task-Oriented Connectivity for Networked Robotics with Generative AI and Semantic Communications}

\author{\IEEEauthorblockN{
Peizheng Li,
Adnan Aijaz
}\\ 
\IEEEauthorblockA{
Bristol Research and Innovation Laboratory, Toshiba Europe Ltd., U.K.\\
Email: {\{peizheng.li, adnan.aijaz\}@toshiba-bril.com}
}
}

\maketitle

\begin{abstract}
The convergence of robotics, advanced communication networks, and artificial intelligence (AI) holds the promise of transforming industries through fully automated and intelligent operations. In this work, we introduce a novel co-working framework for robots that unifies \emph{goal-oriented semantic communication (SemCom)} with a \emph{Generative AI (GenAI)-agent} under a semantic-aware network.
SemCom prioritizes the exchange of meaningful information among robots and the network, thereby reducing overhead and latency. Meanwhile, the GenAI-agent leverages generative AI models to interpret high-level task instructions, allocate resources, and adapt to dynamic changes in both network and robotic environments. This agent-driven paradigm ushers in a new level of autonomy and intelligence, enabling complex tasks of networked robots to be conducted with minimal human intervention.
We validate our approach through a multi-robot anomaly detection use-case simulation, where robots detect, compress, and transmit relevant information for classification. Simulation results confirm that SemCom significantly reduces data traffic while preserving critical semantic details, and the GenAI-agent ensures task coordination and network adaptation. This synergy provides a robust, efficient, and scalable solution for modern industrial environments. 
\end{abstract}


\section{Introduction}
Smart manufacturing and so-called “dark factory” concepts have significantly increased the demand for higher levels of autonomous operation in modern factories. To achieve the desired automation of manufacturing processes, several key technologies are essential: (1) reliable and deterministic communication, (2) automation of assembly lines and robotic operations, 
and (3) real-time information collection, processing, and counteraction generation.

The integration of advanced technologies such as artificial intelligence (AI), machine learning (ML), next-generation communication networks (e.g., B5G/6G), and robotics is expected to revolutionize smart manufacturing. These technologies will enable factories to autonomously complete complex tasks with minimal human intervention~\cite{khang2023enabling}.

Among these advances, robots, and particularly networked multi-robots, form the backbone of modern smart manufacturing. Multi-robot systems are designed to work alongside humans in shared workspaces, boosting productivity, flexibility, and safety. Equipped with advanced communication modules, sensors, actuators, and control systems, multi-robot system can perform a variety of tasks such as assembly, inspection, and material handling. Their ease of programming, operation, and maintenance makes them suitable for a wide range of industrial applications, including mission-critical scenarios, unstructured environments, and tasks subject to continuous change.

Fundamentally, the success of a multi-robot system depends on three interrelated elements: communication, control, and task allocation~\cite{an2023multi}. On the communication front, the deployment of private 5G/6G networks provides reliable, high-speed connectivity for factories and networked robotic nodes~\cite{ghassemian2023standardisation}. On the control side, numerous dedicated schemes have been proposed to enable precise, real-time coordination. By co-designing communication and control, robots can leverage real-time feedback and dynamically adjust control parameters based on current network conditions, thereby minimizing latency, reducing communication overhead, and enhancing overall responsiveness and reliability.

It is worth noting that communication networks themselves are evolving into AI-native intelligent platforms, where AI/ML techniques are embedded to deliver proactive services and applications. In particular, \emph{goal-oriented semantic communication (SemCom)} is emerging as a promising approach for AI-native networks~\cite{strinati2024goal}. Semantic communication focuses on transmitting meaningful content, interpreted via a shared understanding of concepts; while goal-oriented communication emphasizes using that content to achieve specific, shared objectives. Within a robot system, this translates to smarter performance at both the application and connectivity levels. For example, different learning models can be deployed to adapt control and communication methods; robots can access cloud-based resources such as AI models and data analytics to improve decision-making; and the data exchanged can be more selective, reducing overall transmission data volume.

In parallel, the rapid progress in \emph{generative AI} and large language models (LLMs) has demonstrated impressive capabilities for understanding, reasoning, and content generation. These advances have sparked new considerations for applying generative AI in robotics to improve task understanding, decomposition, and allocation~\cite{aristeidou2024generative,ref1}. More recently, the concept of an \emph{AI-agent based on generative AI} (GenAI-agent) has gained traction. Such an agent autonomously perceives its environment, reasons about tasks, and generates context-appropriate outputs or actions. Through continual learning from data and interactions, a GenAI-agent can adapt its behaviour in ways that traditional rule-based systems cannot. 
Generative AI is capable of enhancing multi-robot coordination for complex collaborative tasks, e.g., through real-time information exchange via AI-drive algorithms and synchronization and optimization of collective workflows without human intervention. 

In a networked robotics scenario, the synergy of GenAI-agents, goal-oriented planning, and semantic communication has the potential to yield significant improvements in productivity and efficiency. Accordingly, this paper introduces GenAI-agents and SemCom for future networked robotics, unifying them under a new defined co-working architecture.
By leveraging GenAI-agents, robots can interpret complex instructions, generate adaptive task plans, and handle dynamic environments; simultaneously, SemCom reduces communication overhead and improves the efficiency of multi-robot interactions. Altogether, this architecture enables robots to execute more sophisticated tasks autonomously, enhance human–robot collaboration, and ultimately boost overall efficiency and flexibility in industrial applications.

The main contributions of this paper are as follows:
\begin{itemize}
  \item We integrate \emph{SemCom} into networked robotics to enhance the efficiency of robot-server/robot-robot communication.
  \item We introduce a \emph{GenAI-agent} into multi-robot systems to manage both connectivity resources and robot-related tasks, improving overall autonomy and adaptability.
  \item We present a novel co-working architecture and detail its architecture that combines the \emph{GenAI-agent} with a \emph{semantic-aware open radio access network (Open RAN)}, enabling tight coordination between network and robotic operations.
  \item We validate our approach via a multi-robot anomaly detection use case, demonstrating how SemCom and the GenAI-agent effectively reduce communication overhead while maintaining robust performance.
\end{itemize}

The remainder of this paper is organized as follows. 
Section~\ref{sec:preliminary} presents the background information on networked robotics systems, AI-native SemCom, and the GenAI-agent. 
Section~\ref{sec:synergistic semantic plane} details the proposed architecture for integrating GenAI-agents and semantic-based connectivity into networked robotics. 
Section~\ref{sec:usecase} describes the multi-robot anomaly detection use case, highlighting the benefits of the proposed approach. 
Section~\ref{sec:discussion} discusses open research challenges, 
and Section~\ref{sec:conclusion} concludes the paper.

\section{Preliminary} 
\label{sec:preliminary}

\begin{figure}[t]
    \centering
    \includegraphics[width=0.95\linewidth]{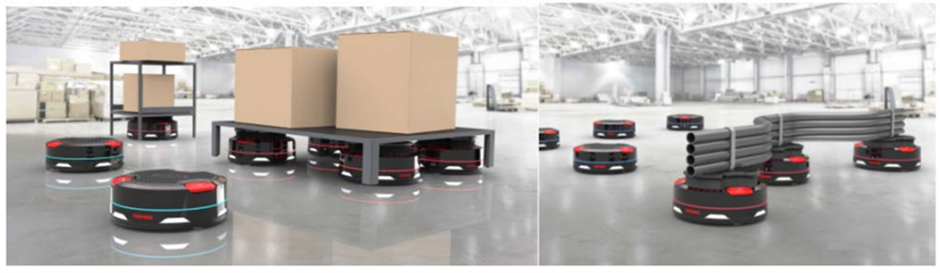}
    \caption{Scenarios of cobot operation for goods moving~\cite{farnham2021umbrella}}
    \vspace{-0.5cm}
    \label{fig:cobot_scenario}
\end{figure}

\subsection{Networked robotics system}
From a communication perspective, interconnecting individual robot nodes via advanced communication protocols facilitates real-time data sharing, remote control, and coordination among multiple robots, thereby paving the way for more intelligent and interconnected industrial ecosystems. Fig.~\ref{fig:cobot_scenario} shows the typical scenarios of multi-robot operation.

A representative example of the multi-robot system is collaborative robots (cobots), such as automated guided vehicles (AGVs) and autonomous mobile robots (AMRs). These systems are equipped with a rich array of advanced sensors (e.g., force sensing, vision systems, and speed and separation monitoring), and leverage robust communication links to navigate autonomously, avoid obstacles, and interact with other robots and humans in shared workspaces.

AGVs and AMRs typically follow predefined rules for paths, behaviours, and tasks. Historically, their design and operation have been discussed from a control-centric perspective~\cite{dehghani2016communication}, which can lead to limitations in handling dynamic operating environments and accurately perceiving complex surroundings.

As an alternative, communication-based control or \emph{communication and control co-design} has been proposed to improve resource allocation and task scheduling~\cite{aijaz2019toward}. In this paradigm, emphasis is placed on the timely exchange of information among different components of a robotic system. By establishing effective communication channels, robots can share data, coordinate their actions, and collaborate on decision-making in real time.

Nonetheless, the operation of cobot systems still faces key challenges, including node synchronization, task allocation, cooperation, and overall communication overhead. In multi-robot scenarios, the frequency of information exchange can be substantial, potentially leading to network congestion, higher latency, and increased energy consumption. Therefore, a pressing question arises: How can a new connectivity paradigm be designed for cobots to optimize task allocation and robot coordination while reducing communication overhead, all without compromising the semantic quality of the transmitted information?

\subsection{AI-native network}
\subsubsection{SemCom}
\emph{SemCom} is an emerging paradigm wherein only the information relevant to achieving a specific goal or task is transmitted. In contrast to traditional communication approaches that focus on maximizing data throughput or minimizing bit-level errors, SemCom targets the exchange of meaningful, task-specific semantic content. By eliminating redundant or irrelevant data, communication becomes more efficient and effective.

SemCom encompasses two complementary concepts. \emph{Semantic communication} focuses on transmitting meaningful content interpreted through a shared understanding of concepts, while \emph{goal-oriented communication} directs that content toward achieving a particular objective. Semantic communication thus provides the interpretive framework, whereas goal-oriented communication ensures that these interpretations collectively contribute to a desired outcome.

\begin{figure*}[t]
    \centering
    \includegraphics[width=0.95\linewidth]{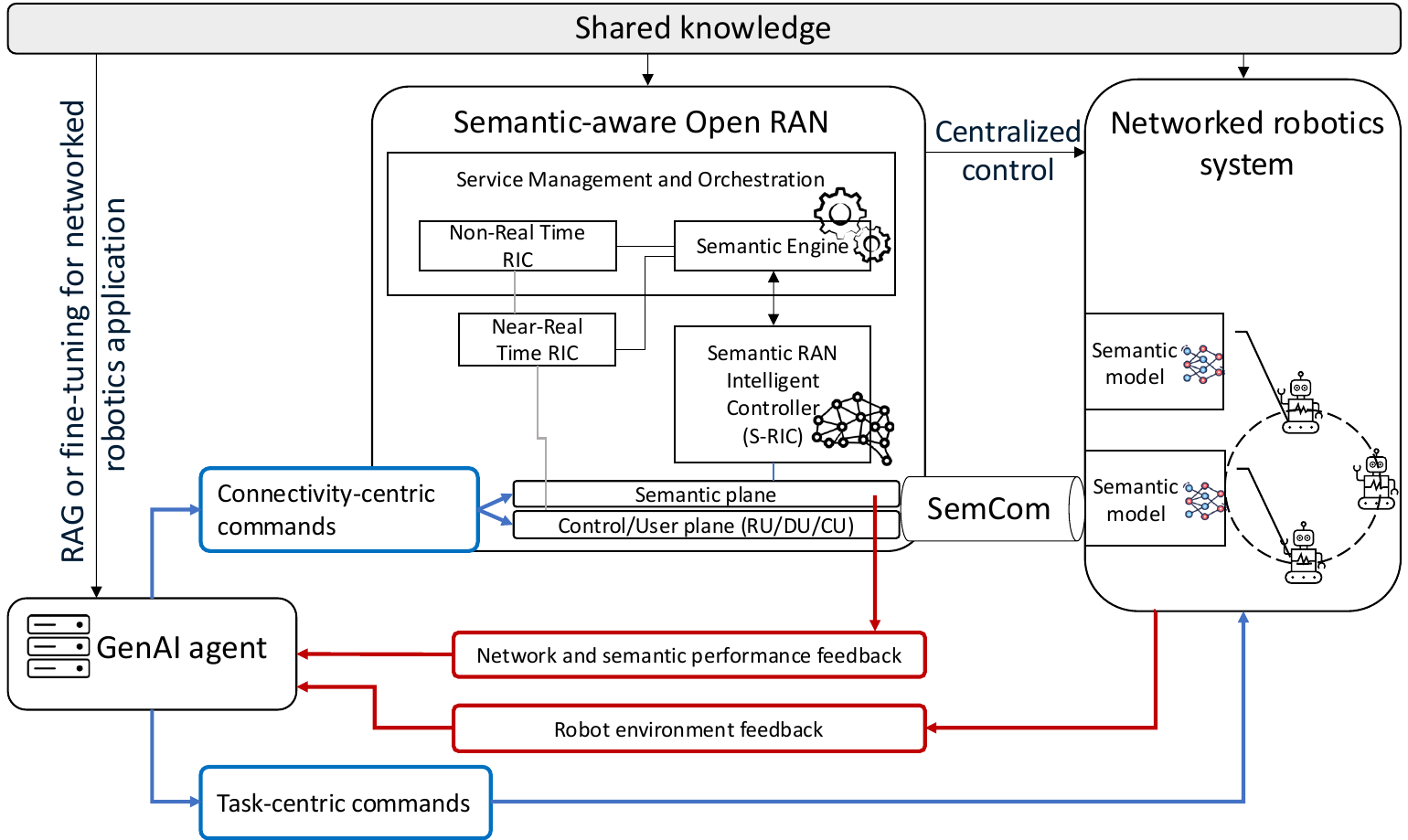}    
    \caption{Illustration of the co-working framework that incorporates GenAI-agent and SemCom for the networked robotics system}
    \label{fig:overall framework}
\end{figure*}

\subsubsection{Semantic-aware connectivity}
Although 6G standardization is still underway, the ITU-R’s “IMT-2030 Framework” includes AI and communication as key pillar features, alongside ubiquitous connectivity and integrated sensing and communication. This aligns with the broader industry and academic exploration of \emph{AI-native networks}, wherein AI is deeply integrated across the network architecture to provide advanced intelligent functions.

Progress toward semantic-aware Open RAN architectures has signalled a crucial step toward AI-native networking. As introduced in the previous works~\cite{strinati2024goal, li2023open}, innovations such as semantic RAN intelligent controllers (S-RIC), semantic plane, and semantic-aware radio/ distributed units try to extend the capabilities of standard Open RAN. By encoding and transmitting only the most relevant semantic content, it becomes possible to reduce data exchange in networked robotics, thereby minimizing overhead without significantly undermining information integrity.

\subsection{GenAI-agent}
Generative AI is a branch of AI dedicated to producing new content, such as images, text, or audio, based on learned patterns in existing data. 
A \emph{GenAI-agent} harnesses generative models to autonomously perceive its connected scenarios, reason about tasks, and produce contextually appropriate actions or outputs. Such an agent can learn from vast quantities of data and interact with its environment to refine its internal models, thus adapting its behavior in real time. By integrating GenAI-agents into multi-robot systems, advanced functionalities like flexible task scheduling, proactive decision-making, and rapid adaptation to new scenarios can be achieved, all while maintaining efficient communication through semantic-aware connectivity.

\section{Synergistic co-working of GAI-agent and SemCom for Robotic System}
\label{sec:synergistic semantic plane}
\subsection{Overall architecture}
Fig.~\ref{fig:overall framework} illustrates the proposed co-working architecture that integrates a GenAI-agent and SemCom within a semantic-aware Open RAN network to support networked multi-robot systems. In this setup, the GenAI-agent, semantic Open RAN, and multi-robot platform share a comprehensive knowledge base related to the vertical application domain. This shared knowledge underpins both training and fine-tuning of semantic models in the Open RAN and robotic subsystems, as well as guiding the GenAI-agent through retrieval-augmented generation (RAG) or fine-tuning. In doing so, each component gains a common understanding of operational contexts, such as manufacturing processes, network conditions, or evolving task requirements.

The core elements of the semantic-aware Open RAN are adapted from our previous work~\cite{strinati2024goal,li2023open}. Specifically, an S-RIC operates in conjunction with non-real-time (non-RT) and near-real-time (near-RT) RICs to orchestrate semantic functionalities across the network. By leveraging the semantic plane, the S-RIC ensures only contextually relevant information flows between network functions and the external robotic system, thereby minimizing unnecessary data transmission. In essence, the Open RAN serves as a central semantic information-processing hub for connected robot nodes, optimizing resource allocation and communication overhead. This transformation, from a conventional data-driven approach to a goal-oriented and meaning-centric paradigm, results in improved efficiency, adaptability, and scalability.

Within this architecture, a GenAI-agent is introduced to enhance autonomous decision-making further. Deployed either as part of the Open RAN setup or alongside it in industrial networks, the GenAI-agent receives constant feedback from robot operations (e.g., progress on task execution) and real-time network metrics (e.g., throughput, capacity, semantic performance). Harnessing advanced generative AI models, the GenAI-agent synthesizes these diverse inputs to generate highly targeted recommendations. The connectivity-centric commands are sent to the semantic-aware Open RAN system, delivering the reconfiguration information to the Semantic/Control/Uesr plane and the central unit (CU), distributed unit (DU), and radio unit (RU), correspondingly. For instance, it can dynamically suggest spectrum usage policies, adapt network slicing strategies, or update semantic models in response to changing conditions. On the robotics side, task-centric commands are generated, wherein GenAI-agent delivers context-aware instructions governing task decomposition, allocation, and execution, thus enabling more precise coordination among multiple robots in complex, and often unpredictable, environments.

\subsection{Workflow of GenAI-agent}
\begin{figure}[t]
    \centering
    \includegraphics[width=0.9\linewidth]{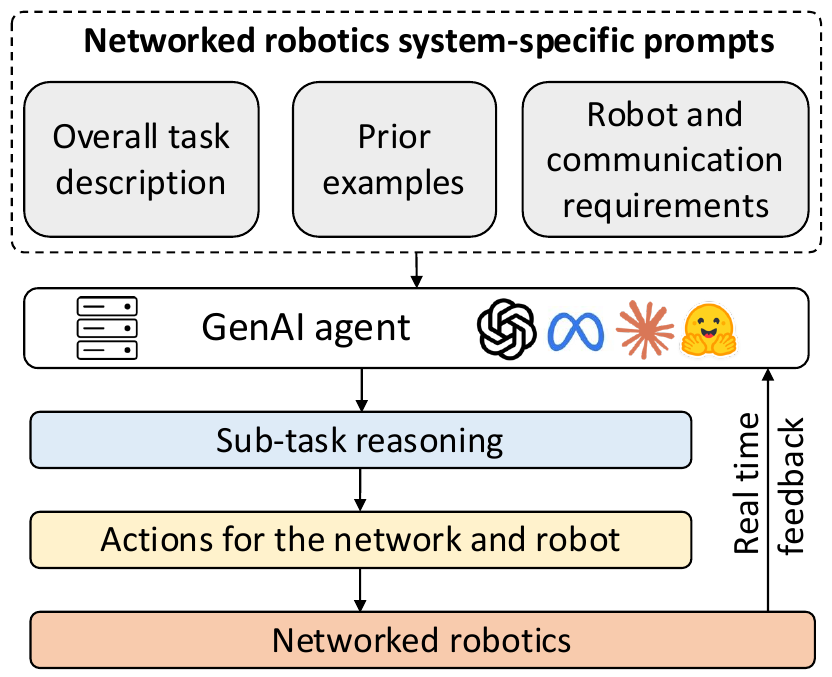}
    \caption{Workflow of the GAI-agent}
    \label{fig:gai-agent architecture}
\end{figure}

Fig.~\ref{fig:gai-agent architecture} illustrates the workflow of the GenAI-agent within a networked robotics control context. Depending on specific application needs, this agent can be built using a variety of generative AI models—including, but not limited to LLMs. GAI services can be accessed via APIs connecting to cloud-based models~\cite{zhou2024large}, or they can be deployed locally for private usage, as discussed in~\cite{li2024large}.

A crucial step in this process is supplying the GenAI-agent with domain-specific knowledge pertaining to the networked robotics. The initial prompts ought to cover relevant information such as the overall task description, prior examples, and both robot- and communication-related requirements. Based on this input, the GenAI-agent decomposes the task into distinct sub-task streams for analysis and reasoning. It then generates corresponding actions for both network and robot components, as described in the preceding sections.

By continuously monitoring network and robotic subsystems, including semantic models, UE traffic, service-level indicators, sensor data, manufacturing requirements, and operational logs, the GenAI-agent seeks to maintain a dynamically optimized state of operation. In turn, the synergy of SemCom and GenAI establishes a robust, responsive, and intelligent industrial infrastructure wherein meaningful insights guide decision-making across distributed components.

\section{Multi-robot system for anomaly detection}
\label{sec:usecase}
\begin{figure}[t]
    \centering
    \includegraphics[width=0.98\linewidth]{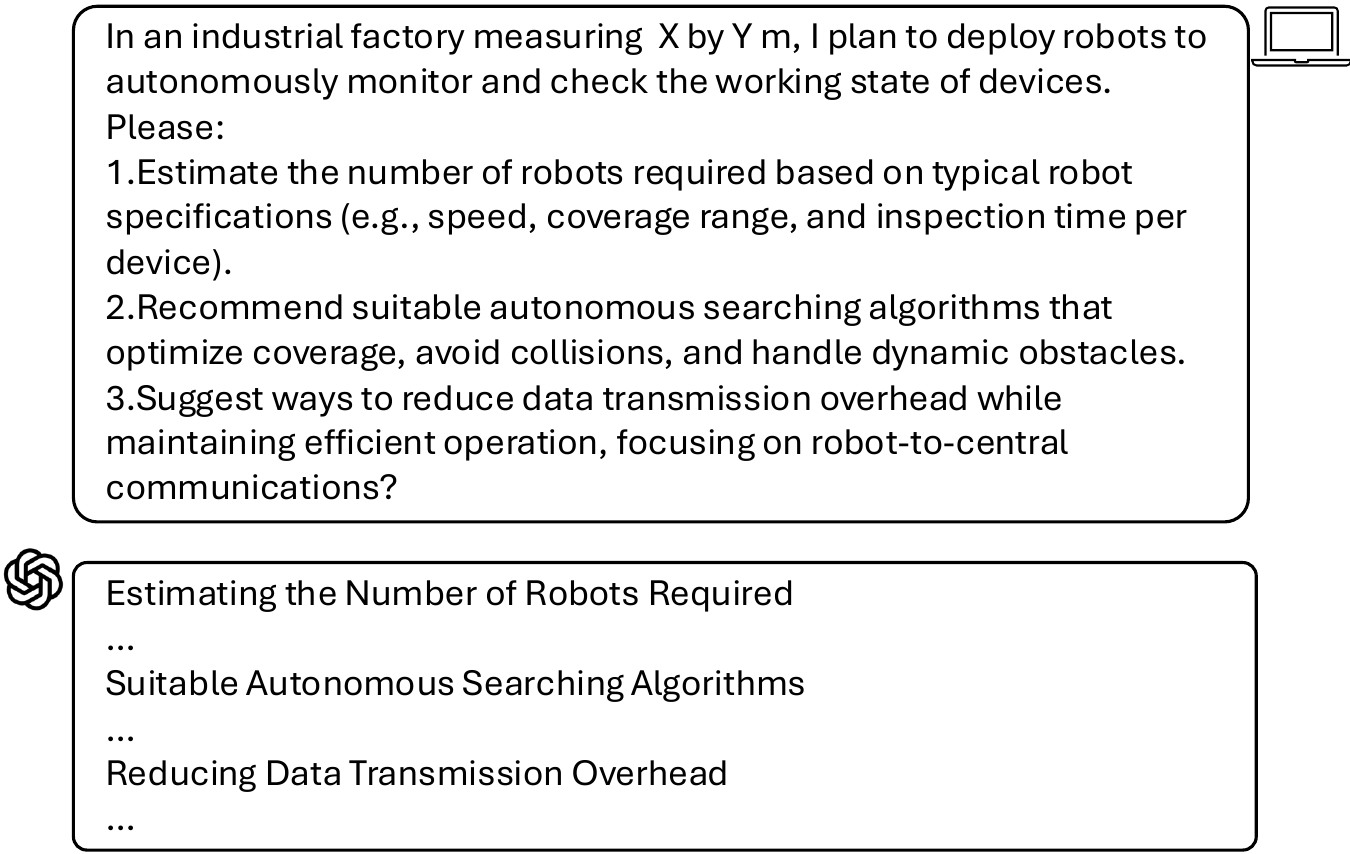}
    \caption{Example of initial prompts to the GAI-agent and its response (partial response omitted for brevity)}
    \label{fig:prompt and response}
\end{figure}

We consider an industrial surveillance scenario in which multiple robots are deployed to detect anomalies (e.g., malfunctioning equipment, safety hazards) in a variety of industrial devices. Fig.~\ref{fig:prompt and response} shows the initial prompts given to the GenAI-agent, along with its responses. These prompts outline key aspects of the robot operating scenario. After a few rounds of dialogue, the GenAI-agent suggests an optimal number of robots, an autonomous search strategy, and a communication-efficient data transmission scheme. We incorporate these recommendations in our subsequent simulation.

\subsection{Use-case model}
Each robot autonomously navigates the environment, captures images within its field of view, and transmits relevant data to a central server over the Open RAN network for anomaly detection and classification. Upon analyzing these images, the central server alerts operators or maintenance personnel if abnormal conditions are detected.

\begin{figure}[t]
    \centering
    \includegraphics[width=0.95\linewidth]{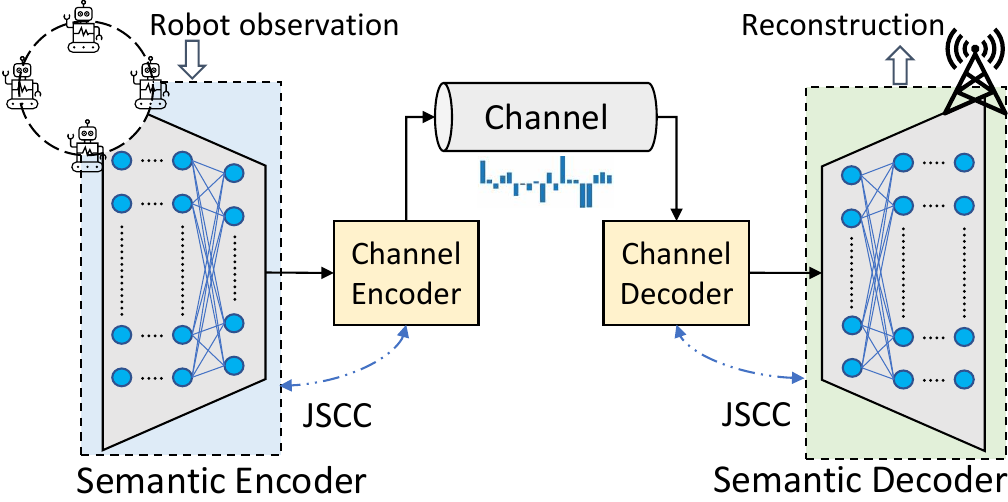}
    \caption{Illustration of VAE used for robot observation reconstruction}
    \label{fig:vae architecture}
\end{figure}

Each robot is assumed to be equipped with a variational autoencoder (VAE)-based semantic encoder for image compression, while the central server hosts the decoder for image reconstruction, as illustrated in Fig.~\ref{fig:vae architecture}, where the histogram below the channel represents the compressed information, and the joint source-channel coding (JSSC)~\cite{bourtsoulatze2019deep} is adopted.

The simulation environment measures 100\,m\,$\times$\,100\,m and contains industrial devices represented by MNIST images (digits 0--9). We define the \textit{abnormal devices} using the even-digit images (0, 2, 4, 6, 8) indicating equipment faults, and odd-digit images (1, 3, 5, 7, 9) are used to indicating functioning \textit{normal devices}.
Obstacles, modelled as walls, are strategically placed to create realistic navigation challenges. Each robot must plan optimal collision-free paths around these walls to reach target devices. The simulation environment with walls and robot nodes is illustrated in 
Fig.~\ref{fig:simulation environment}.

\subsection{Robotics agents}
In this simulation, it is assumed that autonomous robots are deployed in the environment with the capabilities of \textit{1)} sensors: A camera that captures images of nearby devices (within a 2\,m radius); \textit{2)} navigation: The A* path-planning algorithm for obstacle avoidance and route optimization; \textit{3)} communication: A 5G wireless module to exchange data with the central server.

\subsubsection{Communication framework}
We adopt SemCom principles to transmit only the essential information required for classification. Specifically, each robot utilizes a VAE-based encoder to compress the captured images into a low-dimensional latent space vector. The central server, which hosts the VAE decoder, subsequently reconstructs these compressed images. We model the wireless channel using a 3GPP-based indoor path loss model~\cite{3gpp_tr38901} and assume a single antenna link between each robot and the server. The simulation computes path loss, signal-to-interference \& noise ratio( SINR), and maps it to a channel quality indicator (CQI) value, which is then used to determine the modulation and coding scheme (MCS). This helps us estimate the data rate and transmission overhead with higher fidelity, allowing for a realistic evaluation of network performance and the benefits of semantic compression.

\subsubsection{Operational scenarios}
We compare two modes of wireless transmission in this simulation:
\begin{itemize}
  \item \textbf{Branch A (SemCom Enabled)}: Each image is compressed via the VAE encoder into a 20-dimensional latent vector before transmission.
  \item \textbf{Branch B (No Compression)}: Raw images (28\,$\times$\,28 pixels) are transmitted directly, serving as a benchmark to gauge the benefits of semantic compression.
\end{itemize}

\begin{figure}[t]   
    \subfloat[\label{fig:roc}]{
      \begin{minipage}[t]{0.48\linewidth}
        \centering 
        \includegraphics[width=1.75in]{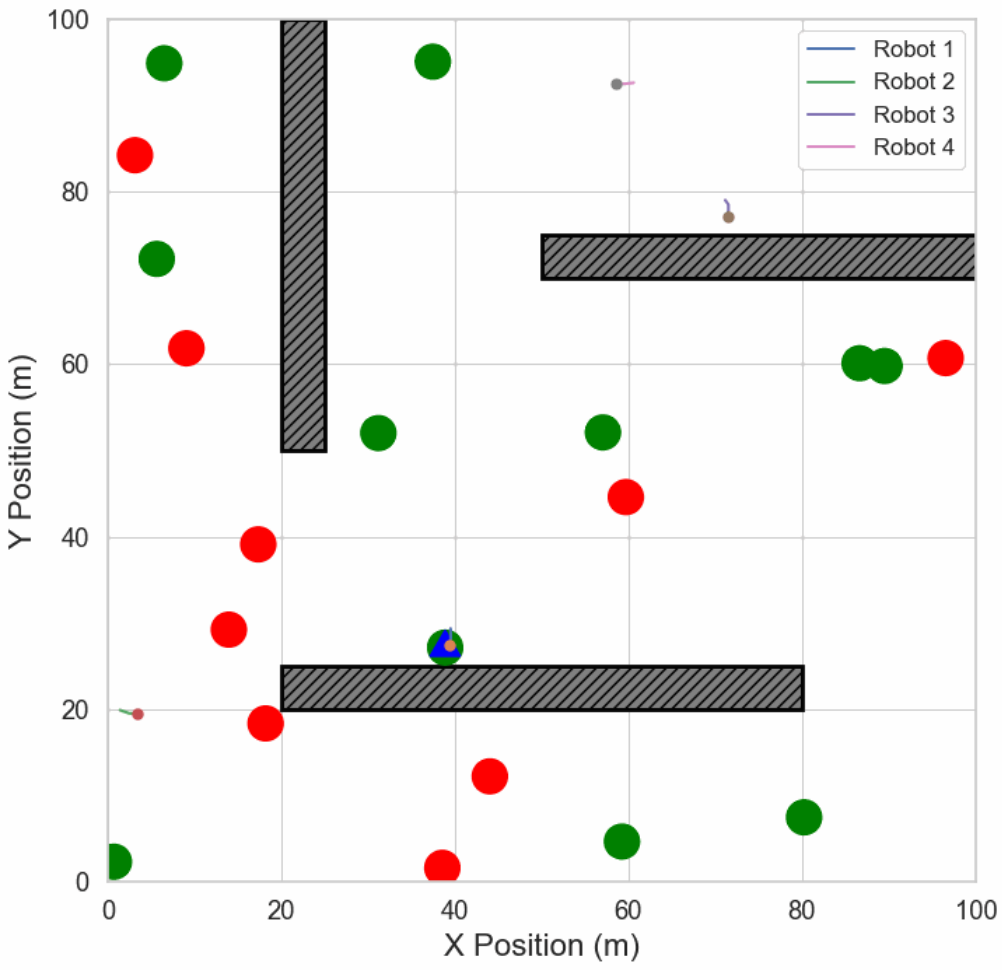} 
        \label{subfig_a}
      \end{minipage}%
      }
        \subfloat[\label{fig:tSNE}]{
      \begin{minipage}[t]{0.48\linewidth}   
        \centering   
        \includegraphics[width=1.75in]{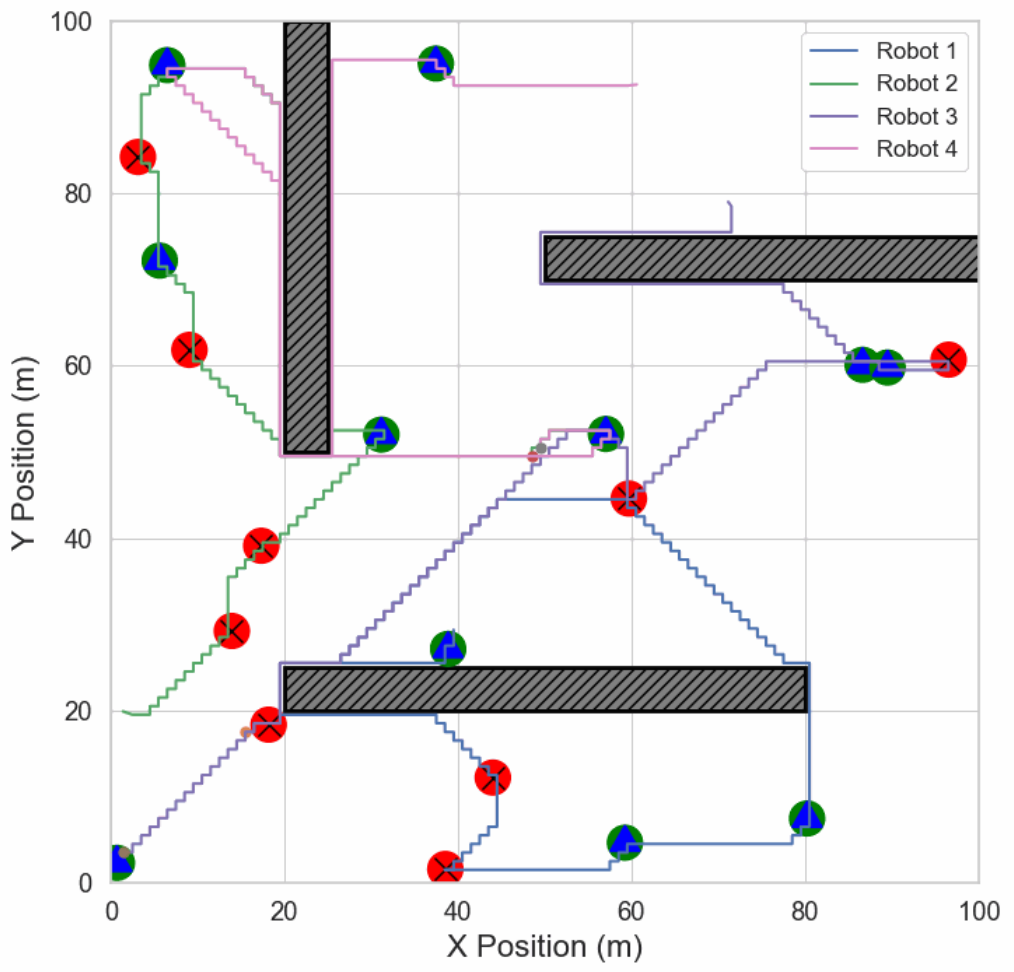}
        \label{subfig_b}
      \end{minipage} 
      }
      \vspace{-1mm}
      \caption{Illustration of simulation environment with 4 robots and 20 devices. (a) shows the initialised environment and robot/device location when the simulation starts. (b) shows the robot's trajectory when the simulation ends.
      } \label{fig:simulation environment}
\vspace{-1.45mm}
\end{figure} 

\subsubsection{Central server functions}
A central processor receives and reconstructs images (if compression is used), then employs a pre-trained MNIST classifier to determine whether each device is normal or abnormal. This processor also manages robot task allocation, directing each robot to undetected devices based on proximity and availability.
\begin{figure}[t]
    \centering
    \includegraphics[width=0.95\linewidth]{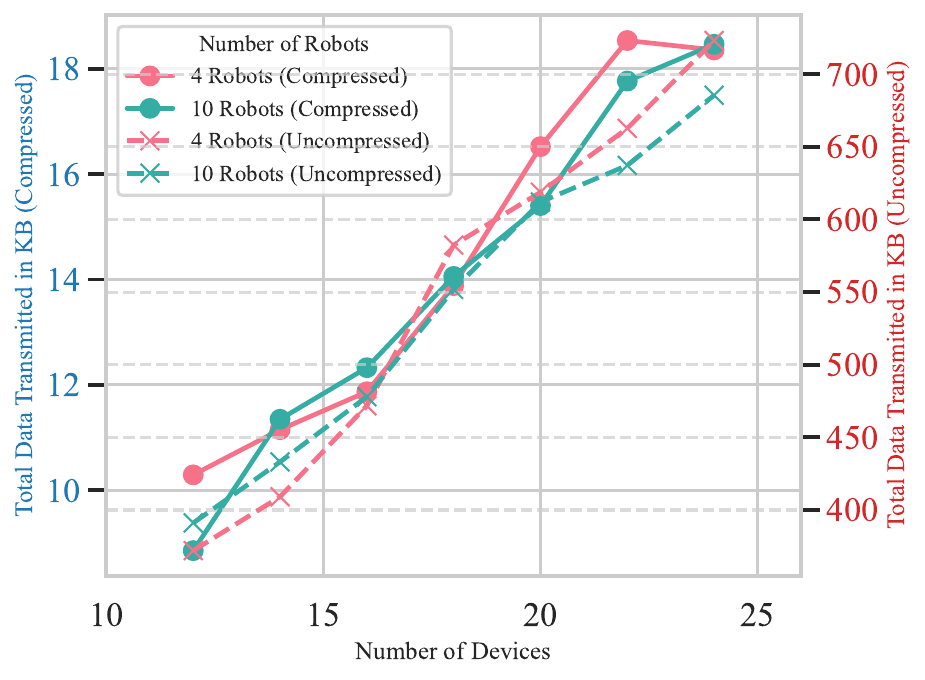}
    \vspace{-0.2cm}
    \caption{Transmitted data comparison for compressed/non-compressed mode.}
    \label{fig:comparison results}
\end{figure}
\subsubsection{Simulation parameters}
Key simulation settings include: total run time is 400\,s, with discrete 1\,s time steps; robot speed set to 2\,m/s (neglecting Doppler effects). 
In this simulation, different numbers of robot nodes and devices are initialised in each simulation round, and we record total transmitted bits over the course of the simulation.
Fig.~\ref{fig:simulation environment} shows the established simulation environment with 4 robots and 20 devices, wherein Fig.~\ref{subfig_a} shows the initialised environment and robot/device locations at the start of the simulation, and Fig.~\ref{subfig_b} demonstrates the overall robot's trajectory during the simulation process.

\subsection{Results}
Throughout the simulation, the system monitors the total volume of data transmitted by the robots. As illustrated in Fig.~\ref{fig:comparison results}, adopting SemCom (Branch A) dramatically reduces the amount of data sent compared with transmitting raw images (Branch B). Moreover, as the number of devices increases, SemCom can shrink data volumes to as little as one-fortieth of the original size while preserving sufficient semantic information for accurate digit-based classification. This significant reduction not only alleviates network load but also enables faster data transfers, both of which are critical for real-time monitoring in industrial environments.

\section{Discussions}
\label{sec:discussion}
\subsection{Acquisition of the knowledge base}
A shared knowledge base is a foundational requirement for the joint use of a GenAI-agent and SemCom in networked robotics. In essence, this knowledge base is a curated dataset containing domain-specific insights relevant to the robots and their environment. For industrial robots, however, publicly available datasets are scarce, and they lack complicated and specific industrial information. That hinders the large-scale adoption of GAI-agent in vertical applications. To address this gap, industrial stakeholders should collect and structure domain-specific data for SemCom model training and GenAI-agent tuning.

\subsection{Implementation of the semantic-aware Open RAN}
SemCom offers a powerful communication paradigm for extracting and transmitting only the most relevant information. Meanwhile, semantic-aware Open RAN aims to integrate SemCom across all layers of the RAN. Despite these benefits, the practical deployment of such an architecture is still in its infancy. Standardized APIs for semantic functions, along with semantic management loops for involved models, should be further refined to ensure robust interoperability and performance in real-world environments.

\subsection{Tuning of the GenAI-agent}
Introducing GenAI-agent into networked robotics systems jointly with semantic connectivity is a crucial step toward autonomous operations. This agent needs to possess a comprehensive and continually updated understanding of the robot environment to effectively handle complex tasks. Beyond RAG and fine-tuning methods, real-world feedback from human operators can be incorporated through reinforcement learning loops, enhancing the agent’s adaptability in rapidly evolving scenarios.

\subsection{Toward real-world integration}
In the current study, the GenAI agent is primarily leveraged at the outset of the simulation to offer high-level insights into multi-robot operations within an indoor industrial environment, while the SemCom model on each robot node focuses on data compression. As a result, the two components, GenAI-agent and SemCom connectivity, are not yet integrated to provide real-time, in-network, or in-robot resource and task allocation. Moving forward, we plan to deploy a practical GenAI agent in a real-world cobot environment to fully assess the synergy between GenAI-agent-driven optimization and SemCom-based connectivity.

\section{Conclusion}
\label{sec:conclusion}
This paper proposes a novel co-working framework that integrates semantic connectivity with a GenAI agent for robotic systems. The approach offers an efficient and adaptive solution for networked multi-robot environments, paving the way for autonomous operations. Through a multi-robot anomaly detection use-case, we demonstrate how semantic connectivity significantly reduces data overhead, while the GenAI agent dynamically manages resources and coordinates tasks. This synergy underscores the potential for reducing operational complexity, improving real-time responsiveness, and scaling automation in next-generation industrial settings.

\section*{Acknowledgment}
This work was supported by the 6G-GOALS project under the 6G SNS-JU Horizon program, n.101139232.
\bibliographystyle{IEEEtran} %
\bibliography{IEEEabrv,references}

\end{document}